\documentclass[letterpaper, 10pt, journal]{IEEEtran}

\usepackage{xcolor,soul,framed} 

\colorlet{shadecolor}{yellow}
\usepackage[pdftex]{graphicx}
\graphicspath{{../pdf/}{../jpeg/}}
\DeclareGraphicsExtensions{.pdf,.jpeg,.png}

\usepackage[cmex10]{amsmath}
\usepackage{array}
\usepackage{mdwmath}
\usepackage{mdwtab}
\usepackage{eqparbox}
\usepackage{url}
\usepackage{cite}
\usepackage{hyperref}

\usepackage{float}     

\usepackage{ragged2e} 
\usepackage{booktabs,makecell, multirow, tabularx}
\usepackage[ruled,linesnumbered]{algorithm2e}
\usepackage[numbers]{natbib}
\SetKwComment{Comment}{// }{}

\begin{document}


\title{\LARGE \bf
A Vehicle-Infrastructure Multi-layer Cooperative Decision-making Framework
}

\author{Yiming Cui$^{1}$, Shiyu Fang$^{1}$, Peng Hang$^{1}$, ~\IEEEmembership{Senior Member,~IEEE,} and Jian Sun$^{1}$
\thanks{*This work was supported in part by the State Key Laboratory of Intelligent Vehicle Safety Technology under Project No. IVSTSKL-202414, the National Natural Science Foundation of China (52472451), the Shanghai Scientific Innovation Foundation (No.23DZ1203400), and the Fundamental Research Funds for the Central Universities.}
\thanks{$^{1}$Y. Cui, S. Fang, P. Hang and J. Sun are with the State Key Laboratory of Intelligent Vehicle Safety Technology and the College of Transportation, Tongji University, Shanghai 201804, China.
        {\tt\small \{2310796, 2111219, hangpeng, sunjian\}@tongji.edu.cn}}%
\thanks{Corresponding author: Peng Hang}
}

\maketitle

\vspace{-0.8cm}
\begin{abstract} 
Autonomous driving has entered the testing phase, but due to the limited decision-making capabilities of individual vehicle algorithms, safety and efficiency issues have become more apparent in complex scenarios. With the advancement of connected communication technologies, autonomous vehicles equipped with connectivity can leverage vehicle-to-vehicle (V2V) and vehicle-to-infrastructure (V2I) communications, offering a potential solution to the decision-making challenges from individual vehicle's perspective. We propose a multi-level vehicle-infrastructure cooperative decision-making framework for complex conflict scenarios at unsignalized intersections. First, based on vehicle states, we define a method for quantifying vehicle impacts and their propagation relationships, using accumulated impact to group vehicles through motif-based graph clustering. Next, within and between vehicle groups, a pass order negotiation process based on Large Language Models (LLM) is employed to determine the vehicle passage order, resulting in planned vehicle actions. Simulation results from ablation experiments show that our approach reduces negotiation complexity and ensures safer, more efficient vehicle passage at intersections, aligning with natural decision-making logic.
\end{abstract}


\section{Introduction}
\begin{figure*}[htbp]
  \begin{center}
  \centerline{\includegraphics[width=7.16in]{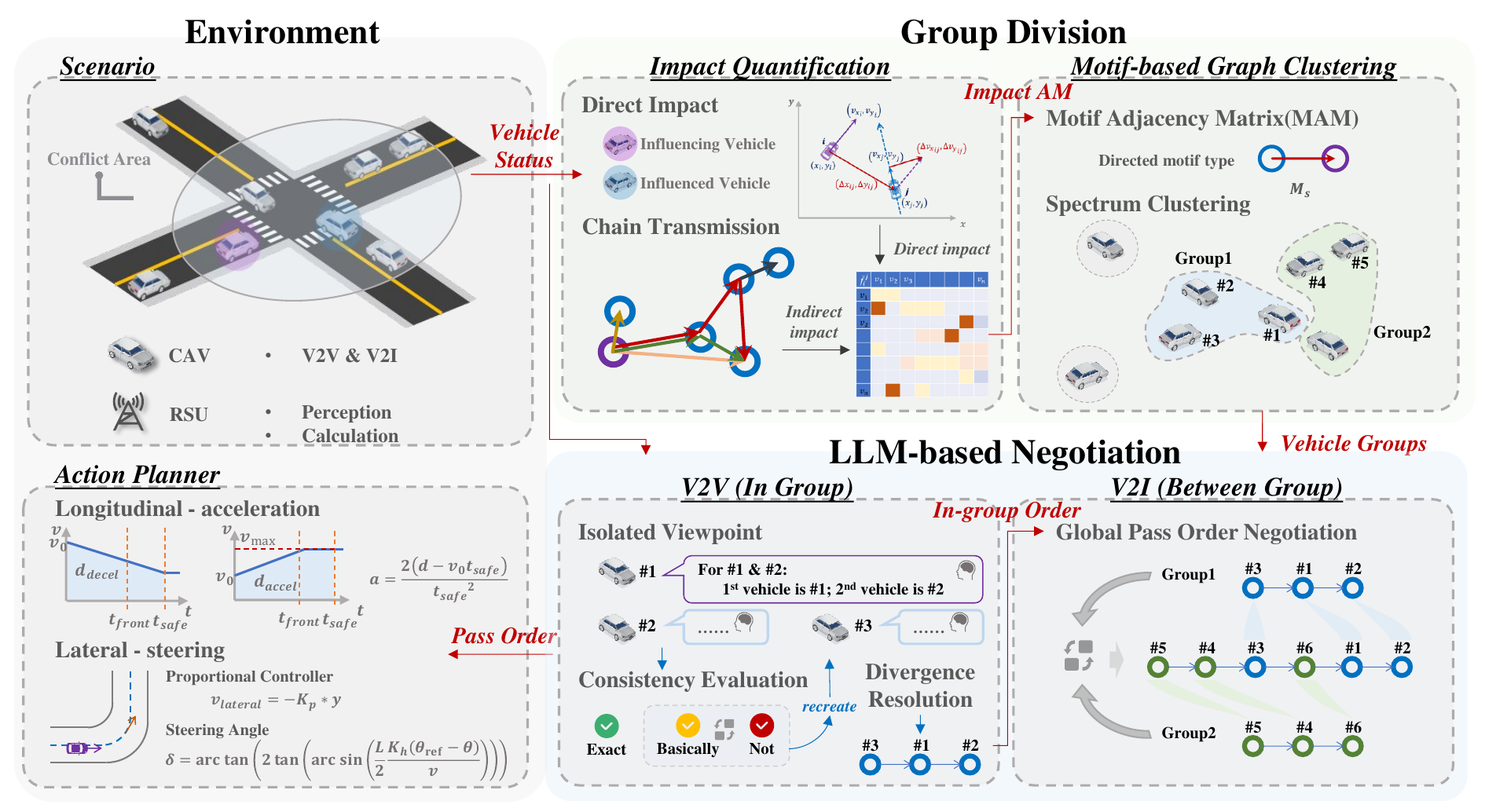}}
  \caption{Overview of vehicle-infrastructure multi-layer cooperative decision-making framework for CAVs at unsignalized intersection.}\label{fig_framework}
  \end{center}
  \vspace{-0.8cm}
\end{figure*}

Currently, autonomous vehicles have moved into field testing and practical application stages worldwide \cite{hu2023autonomous, ali2024investigating}. However, during the trial phase, safety and efficiency issues arise in complex scenarios \cite{schwall2020waymo} such as intersections, where the limited decision-making capability of single-vehicle autonomous driving algorithms leads to decision failures. 
With the advancement of connected communication technologies, autonomous vehicles equipped with connectivity capabilities can leverage V2V, V2I communications and cooperative effort, offering a potential solution to the limitations associated with decision-making from the perspective of individual vehicles.

In consideration of the advantages in improving safety, efficiency, and reducing energy consumption through the coordination of vehicles and road infrastructure \cite{li2021modeling}, cooperative approaches have become a major focus of research in autonomous driving. Existing studies on cooperative decision-making primarily emphasize V2V cooperation and V2I cooperation.
Cooperative decision-making methods for V2V communication can be categorized into rule-based methods \cite{dresner2008multiagent, xiao2021rule}, optimization-based methods \cite{pan2022convex, sun2020cooperative}, game theory-based methods \cite{hang2020human, wang2021competitive} and learning-based methods \cite{zhou2022multi, oroojlooy2023review}. However, due to the limitations in the perception and computational capabilities of individual CAV, first, it is difficult to effectively address the issue of large-scale V2V cooperation from the perspective of the vehicle, for which it is necessary to decompose the large-scale problem and consider partitioning the whole into cooperative vehicle groups. Second, the outcomes of V2V cooperation may lead to local optimal solutions, resulting in infeasible outcomes or negative impact for overall traffic system, which means a higher-level decision role is necessary from the perspective of road and infrastructure.

Due to the existence of autonomous vehicles with varying levels of intelligence, there are differences in their autonomous decision-making capabilities \cite{hu2022review}. Additionally, equipping each autonomous vehicle with sufficient computational power is both costly and impractical. Therefore, vehicle-to-infrastructure (V2I) and vehicle-to-infrastructure-cloud (V2I-C) cooperation are considered more practical collaborative approaches \cite{suo2023proof}. Currently, most research on V2I cooperation focuses on the development of Intelligent Transportation Systems (ITS) from a macro perspective of traffic management and control. For example, after collecting or receiving vehicle status information from roadside units (RSUs), intelligent signal control \cite{sun2022eco, bai2022hybrid}, dynamic speed and lane management \cite{khondaker2015variable, pasha2024dynamic}, and collision warning \cite{sanguesa2016survey} can be implemented. Some studies also explore scenarios where roadside infrastructure, as the owner of global vehicles information, participates in the decision-making process of autonomous vehicles, contributing to their planning and action \cite{li2024v2x}. However, this macro-level V2I cooperation overlooks the individualized needs of each vehicle, and such implicit requirements are often difficult to be explicitly expressed and comprehend, leading to their exclusion from decision-making processes.

Given the extensive training data and large model parameters, Large Language Models (LLMs) possess strong comprehension and reasoning abilities, and their rapid development in recent years has provided new solutions for collaborative driving.
Current studies have shown that LLMs exhibit great potential in the decision-making tasks of individual autonomous vehicles \cite{fu2024drive, wen2023dilu}. At the same time, research in other fields has also explored the use of LLMs for modeling and solving multi-agent decision-making problems \cite{wang2024survey, ni2024mechagents}. 
Building upon the aforementioned research, the integration of LLMs into cooperative decision-making can effectively simulate the decision logic of individual vehicles and human drivers. Considering the characteristics of autonomous driving and the modeling and solving requirements of cooperative decision-making, this approach holds promise for achieving safe and efficient cooperative driving. However, current efforts with LLM primarily focus on individual vehicle or simple V2V scenarios, whereas achieving the goal of cooperative driving requires the joint effort of both V2V and V2I, which remains largely unexplored.

To address the aforementioned challenges, we propose a multi-level vehicle-infrastructure cooperative decision-making framework for complex conflict scenarios at unsignalized intersections as shown in Fig.~\ref{fig_framework}. 
Firstly, the vehicle impact quantification and transmission relationships are defined based on vehicle states, with cumulative impact serving as the basis for clustering cooperative vehicle groups using motif-based graph clustering. Subsequently, within and between these vehicle groups, LLM-based methods are employed to facilitate vehicle pass order negotiations, and the resulting decisions are used to generate motion plans, thereby achieving defusion of V2V and V2I cooperation.
The contributions of this study are as follows:  
\begin{itemize}
    \item A feasible multi-level cooperative decision-making framework is proposed for integrating vehicle-to-vehicle (V2V) and vehicle-to-infrastructure (V2I) cooperation.
    \item A method based on direct "influence quantification – cumulative influence – vehicle group division" is employed to determine V2V cooperative groups, which better accounts for vehicle impact and transmission relationships, thereby simplifying the scale of V2V cooperation problem.
    \item A negotiation and decision-making method based on LLM is introduced for intra-group and inter-group coordination to simulate the negotiation process between vehicles and vehicle groups, aligning more closely with natural decision-making logic for intersection passage.
\end{itemize}


\section{Problem Description}
\subsection{Scenario Description}
The proposed method aims to address the problem of enabling autonomous vehicle groups to pass through complex conflict areas in a reasonable and safe manner. As a typical complex conflict scenario, unsignalized intersections are critical hotspots for frequent accidents and common areas where autonomous driving interactions breakdown. Therefore, they have been selected as the focus of this study, as illustrated in Fig.~\ref{fig_framework}-\textit{Scenario}. To better describe the problem, the following assumptions are made for vehicles and roadside infrastructure.

To provide a clearer description of the problem, the following assumptions are made: communication between vehicles and roadside infrastructure is assumed to be delay-free, and perception is assumed to be error-free. Moreover, the research subjects are defined as all Connected Automated Vehicles (CAVs), equipped with Vehicle-to-Vehicle (V2V) and Vehicle-to-Infrastructure (V2I) communication capabilities.


The scenario involves \(N\) vehicles (\(\mathcal{C} = \{c_1, c_2, \ldots, c_N\}\)) approaching the conflict area, as well as the upstream regions that are taken into consideration, such as intersections. Vehicles arrive from opposite directions at each entry point. The state variable of vehicle \(c_i\) is denoted as \(\mathbf{s}_i = \left(\left(x_i, y_i\right),  \left(v_{xi}, v_{yi}\right)\right)\), which stores its two-dimensional position and velocity respectively.

\subsection{Formulating the Passage Negotiation Decision Problem}
For larger-scale vehicles experiencing spatiotemporal conflicts at intersections, it is necessary to determine a feasible pass order within the conflict area. The process, starting from vehicle states and proceeding through vehicle group division, intra-group pass order negotiation, and inter-group (global) pass order negotiation, can be formulated as follows:
\begin{equation}
    \begin{aligned}
        & G = \{G_1, G_2, \cdots, G_m\} = f_1\left( \textbf{s}_1, \textbf{s}_2, \cdots, \textbf{s}_N \right)\\
        & \bigcup_{j=1}^m G_j = \mathcal{C} \quad \text{and} \quad G_i \cap G_j = \emptyset \, (i \neq j) \\
        & O_{in}\left(G_k\right) = f_2\left(ov_{c_{k1}}, ov_{c_{k2}}, \cdots, ov_{c_{kn}}\right), c_{ki} \in G_k\\
        & O_{all} = f_3\left( O_{in}\left(G_1\right), O_{in}\left(G_2\right), \cdots, O_{in}\left(G_m\right) \right)\\
    \end{aligned}
    \label{eq_formulation}
\end{equation}
where \( G = \{G_1, G_2, \cdots, G_m\} \) represents the partitioning of all vehicles in the scenario into \( m \) mutually exclusive vehicle groups based on their states, through the function \( f_1(\cdot) \). For each vehicle group \( G_k \), the intra-group pass order \( O_{in}(G_k) \) is determined by the pass order preferences \( ov_{c_{ki}} \) of the vehicles within the group, through the function \( f_2(\cdot) \). Finally, the overall global pass order \( O_{all} \) is derived by synthesizing the intra-group pass orders \( O_{in}(\cdot) \) from each group, using the function \( f_3(\cdot) \).  

\section{Methodology}
This chapter introduces the proposed multi-level vehicle-infrastructure cooperative decision-making framework. First, the motif-based spectral clustering approach for cooperative group division is presented. Then, the focus shifts to the implementation and specific design of the decision-making process based on LLMs for negotiating the pass order, both within the cooperative vehicle group and between different groups. Finally, a brief explanation is provided on how vehicle actions are computed once the traffic order is determined.

\subsection{Group Division}
An important issue with centralized solving methods is that the computational complexity increases exponentially as the number of vehicles to be processed grows. Therefore, it is necessary to design a method that divides vehicles in the vicinity of conflict areas into several cooperative groups, thereby simplifying the computational complexity and improving the ability to reach a consensus in negotiation.

\begin{figure}[htbp]
  \begin{center}
  \centerline{\includegraphics[width=3.5in]{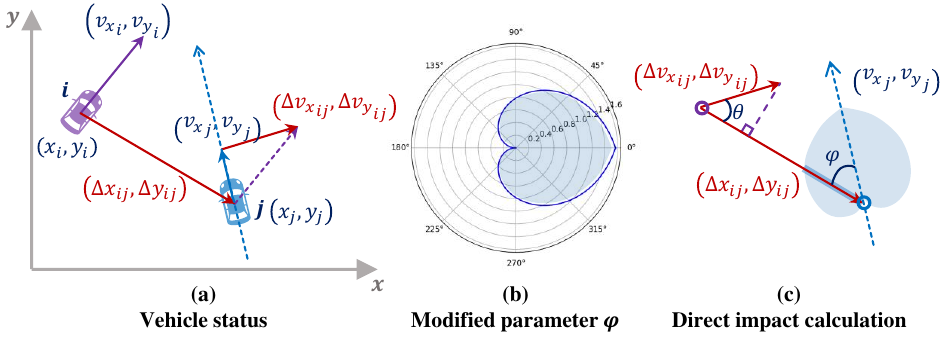}}
  \caption{Direct impact calculation between any two vehicles.}\label{fig_impact}
  \end{center}
  \vspace{-0.8cm}
\end{figure}

To achieve vehicle group division, a natural thought is to determine the mutual influence relationships between vehicles as the basis for group partitioning. First, we define a method for quantifying and calculating vehicle direct influence. For two vehicles in the Cartesian coordinate system, based on their current states (two-dimensional position and velocity), the direct influence of vehicle \(i\) on vehicle \(j\) is calculated using the formula shown in Eq.~\ref{eq_direct_influence}.
\begin{equation}
    \begin{aligned}
        & \theta_{ij} = \arccos \left( \frac{(\Delta v_{xij}, \Delta v_{yij}) \cdot (\Delta x_{ij}, \Delta y_{ij})}
        {\| (\Delta v_{xij}, \Delta v_{yij}) \| \| (\Delta x_{ij}, \Delta y_{ij}) \|} \right) \\ 
        & \varphi_{ij} = \pi - \arccos \left( \frac{(v_{xj}, v_{yj}) \cdot (\Delta x_{ij}, \Delta y_{ij})}
        {\| (v_{xj}, v_{yj}) \| \| (\Delta x_{ij}, \Delta y_{ij}) \|} \right)\\
        & {f_{i}^{j}}_0 = 
        \left( \frac{\| (\Delta x_{ij}, \Delta y_{ij}) \|}{\| (\Delta v_{xij}, \Delta v_{yij}) \| \cos \theta_{ij}} \right)^{-1}\\
        & f_{i}^{j} = {f_{i}^{j}}_0 \times \frac{\sin\left(\pi - \varphi\right)}{2} \\
        & H(x) = 
            \begin{cases} 
            0, & \text{if } x < 0 \\
            1, & \text{if } x \geq 0
            \end{cases} \\
        & \tilde{f}_{i}^{j} = f_{i}^{j} \times H(f_{i}^{j})\\
    \end{aligned}
    \label{eq_direct_influence}
\end{equation}
where \(i\) be the vehicle exerting influence, and \(j\) the affected vehicle. \((v_{xj}, v_{yj})\) are the velocity components of vehicle \(j\), and \((\Delta x_{ij}, \Delta y_{ij})\) are the relative positions of vehicle \(j\) to vehicle \(i\). \((\Delta v_{xij}, \Delta v_{yij})\) are the relative velocities, representing how fast vehicle \(i\) approaches vehicle \(j\). \(\theta_{ij}\) is the angle between the relative position and velocity vectors. \(\|\mathbf{\cdot}\|\) denotes the vector magnitude. The influence quantification result \({f_{i}^{j}}_0\) is the inverse of the time it takes for vehicle \(i\) to reach vehicle \(j\), representing how closely vehicle \(i\) approaches vehicle \(j\).

At the same time, considering that the influence between vehicles may be anisotropic, with the influence of the leading vehicle being more significant while the effect of the vehicle directly behind is minimal, an additional parameter \(\varphi_{ij}\) is introduced to represent this characteristic. This parameter is the complement of the angle between the velocity direction of vehicle \(j\) and the relative position vector \(\left(\Delta x_{ij}, \Delta y_{ij}\right)\). The base influence \(f_{i}^{j}\) is then corrected using the term \(\frac{\sin\left(\pi - \varphi\right)}{2}\). 
If the calculated value of \({f_{i}^{j}}_0\) is negative, it indicates that the vehicles are traveling in opposite directions without interference. Therefore, a Heaviside function is introduced to assign a value of 0 to negative values, while positive values remain unchanged. The final result of the direct influence is denoted as \(\tilde{f}_{i}^{j}\).

Furthermore, in addition to the direct influence between vehicles, indirect influence also exists and plays a crucial role.
To quantify the direct influence between vehicles and the cumulative influence after the chain transmission of influence, this paper proposes a multi-level influence calculation method based on the adjacency matrix and path search. Through recursive traversal of the adjacency matrix, all possible influence paths from the originating vehicle to other vehicles are extracted. A global search is conducted on all influence paths between vehicles to construct a complete set of influence links. Based on the global path set, all influence paths from the influencing vehicle to the target influenced vehicle are selected. The influence value on each path is accumulated through the product of edge weights along the path, reflecting the attenuation or amplification effects of influence during the chain transmission. The total accumulated influence of vehicle \(i\) on vehicle \(j\) is the sum of the influence values of all paths. The key formula in the process is shown in Eq.~\ref{eq_chain_trans}.
\vspace{-0.2cm}
\begin{equation}
    \begin{aligned}
        & A = \left[ \tilde{f}_{i}^{j} \right]_{n} \\
        & A' = \frac{A}{\max(A)}= \left[ {f}_{ij} \right]_{n}\\ 
        & P_{i \to j} = \{ p_k \mid p_k = (v_1, \dots, v_n), \, A'_{v_x, v_{x+1}} > 0, \, v_1 = i, v_n = j \}\\
        & f(p_k) = \prod_{(v_x, v_{x+1}) \in p_k} f_{v_x, v_{x+1}} \\ 
        & F_{i \to j} = \sum_{p_k \in P_{i \to j}} f(p_k) \\ 
    \end{aligned}
    \label{eq_chain_trans}
\end{equation}
where \(A\) represents the direct influence matrix obtained by quantifying the influence between any two vehicles. After normalization, we get \(A'\). \(P_{i \to j}\) is the set of all possible paths for the influence transmission from vehicle \(i\) to vehicle \(j\), and \(f(p_k)\) is the indirect influence of vehicle \(i\) on vehicle \(j\) along the propagation path \(p_k\). \(F_{i \to j}\) is the accumulated influence value for vehicle \(i\) on vehicle \(j\) after the influence is transmitted through all possible paths.

Through the calculation of the accumulated influence between vehicles, we obtain an accumulated influence adjacency matrix, which represents a directed weighted graph. Clearly, based on the definition of direct influence calculation and transmission relationships, the influence between vehicles is asymmetric so that the accumulated influence adjacency matrix is non-symmetric, and the weighted graph is a directed weighted graph. Standard clustering methods, such as standard spectral clustering and minimum cut, are typically applicable only to undirected graphs. Therefore, we adopt a motif-based spectral clustering method to achieve vehicle group division. The Motif Adjacency Matrix (MAM) is a general node similarity metric that extends the traditional adjacency matrix by utilizing higher-order structures (the motifs) in the graph. By incorporating edge direction and weight, it captures more complex relationships between nodes, as defined in Eq.~\ref{eq_motif_AM}.
\begin{equation}
    \begin{aligned}
        & M_{ij} = \frac{1}{|E_M|} \sum_{\sigma \in \mathcal{S}_{\mathcal{M}, \mathcal{A}}^{\sim} } \sum_{\{k_2, \dots, k_{m-1}\} \subseteq V} J_{k, \sigma} G_{k, \sigma}\\ 
        & 
    \end{aligned}
    \label{eq_motif_AM}
    \vspace{-0.5cm}
\end{equation}
where \(M_{ij}\) is the element of the motif adjacency matrix; \(|E_M|\) is the size of the motif's edge set; \(\mathcal{S}_{\mathcal{M}, \mathcal{A}}^{\sim}\) is the anchored automorphism class, representing unique node permutations where a subset of nodes (anchor set \(\mathcal{A}\)) is fixed, and equivalent structures are removed; \(J_{k, \sigma}\) is an indicator function that checks if a motif matches the graph structure under a node mapping \(\sigma\). If the motif matches, \(J_{k, \sigma} = 1\), otherwise \(J_{k, \sigma} = 0\); \(G_{k, \sigma}\) represents the motif's weighted contribution, calculated using the average edge weight in the motif. For more details, refer to the functional motif adjacency matrix in \cite{underwood2020motif}.

The computed MAM is a symmetric matrix, and spectral clustering can be applied to achieve vehicle group partitioning based on this symmetric matrix. In this method, a random walk Laplacian matrix is constructed based on the MAM, the first \(k\) eigenvectors of the Laplacian matrix are extracted, and the k-means++ algorithm is used for clustering. The number of clusters corresponding to the maximum silhouette coefficient is chosen as the optimal number of clusters, and the corresponding clustering result is saved as the final division of the vehicle groups.

\vspace{-0.3cm}
\subsection{Pass Order Negotiation}
LLMs, trained on large-scale corpora, are capable of mastering a vast amount of content related to traffic rules and relevant physical knowledge, taking into account multiple factors such as safety and efficiency in traffic management. Compared to other methods, LLM can clearly explain the reasons behind decisions in natural language, making them more aligned with human decision-making logic. They can also simulate collaborative interactions between vehicles, helping them reach a consensus on the pass order at conflict zone, and are expected to perform better in negotiation tasks. 

To determine the pass order of vehicles in the conflict area, the LLM-based approach is sequentially employed for negotiating the pass order both within vehicle groups and between groups, thus achieving a seamless integration of V2V and V2I cooperation.
First, for each vehicle group after division, the pass order within the group is determined through consensus alignment and divergence resolution negotiation. Furthermore, to ensure that the pass orders of all vehicles are consistent and non-conflicting, a global vehicle pass order is generated through inter-group negotiation, based on the intra-group pass orders that have already been agreed upon.

\begin{figure}[htbp]
  \vspace{-0.3cm}
  \begin{center}
  \centerline{\includegraphics[width=3.5in]{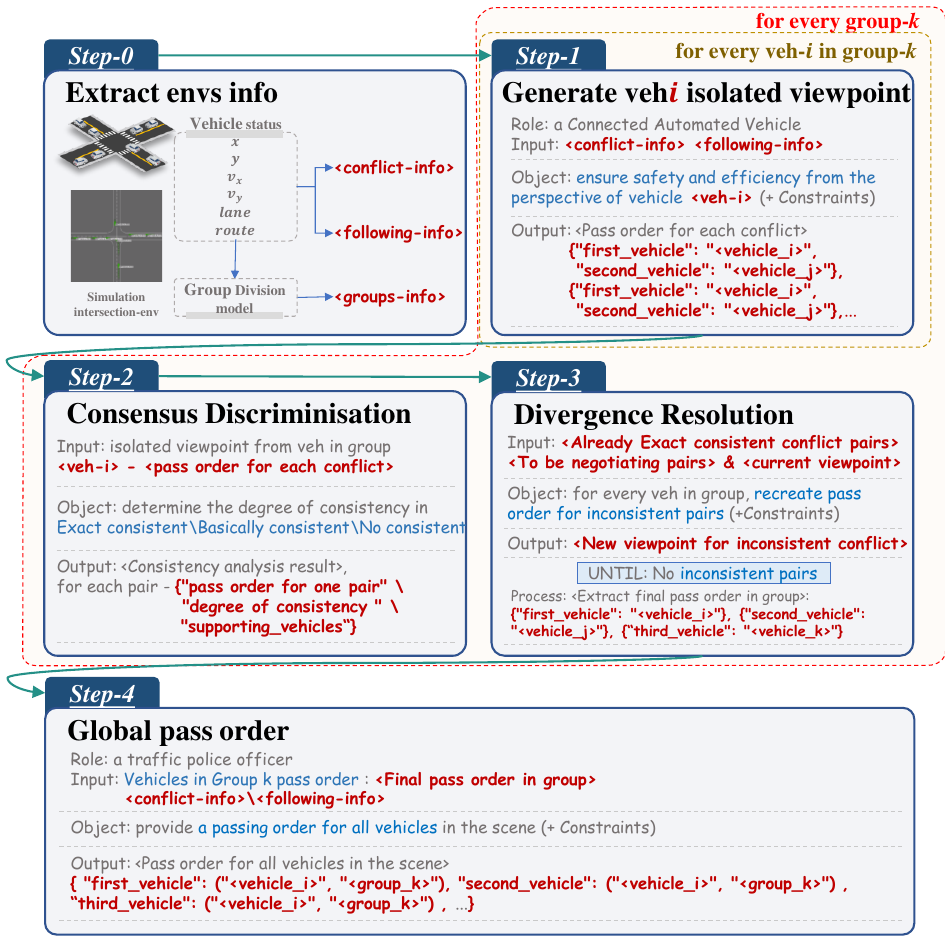}}
  \caption{LLM negotiation implementation process.}\label{fig_LLM}
  \end{center}
  \vspace{-0.8cm}
\end{figure}


\subsubsection{Multi-agent negotiation in group}
First of all, based on the state information of the vehicles within the group, each vehicle in the group independently generates its own viewpoint on the pass order of the vehicles within the group. Starting from the vehicle's own decision-making logic, each vehicle initiates a pass order based on LLM decision-making, considering factors that benefit its own safety and efficiency. The specific expression of this pass order is that for each pair of conflicting vehicles within the group, it is determined which vehicle should pass the conflict point first and which one should pass later.

Then, for each conflict, the pass order viewpoint generated by each vehicle is evaluated to determine the decision on the order of the conflicting vehicles, and it is checked whether the vehicles within the group reach a consensus on the order of these two vehicles. Based on the result, LLM can classify the consistency as \textit{Exact consistent}, \textit{Basically consistent}, or \textit{No consistent}. For exact consistent perspectives, they are retained in the final intra-group pass order negotiation result. For basically consistent and inconsistent results, further divergence resolution is performed using LLM.

Based on the pass order of the conflicting vehicle pairs that have already reached a consensus, the preferences for the pass order in the results of basically consistent and inconsistent vehicles, as well as the conflict and following information of all vehicles within the group, are comprehensively considered. Ultimately, for each conflicting vehicle pair, the consensus result is added to the final intra-group pass order negotiation result.

During the negotiation process, a few basic rules must be strictly adhered to: (1) no vehicles should be omitted or added; (2) a following vehicle with a trailing relationship cannot pass before the leading vehicle; (3) the pass order chain formed by the pairwise pass order decisions within the group must not form a cycle (e.g., A before B, B before C, and C before A is not allowed). If these conditions are not met, further negotiation is required until a feasible result is reached.

\subsubsection{Global negotiation between groups}
Conducting V2V cooperation only among a subset of vehicles without considering other vehicles outside the group may lead to conflicts caused by contradictions in the pass order between groups. Therefore, to ensure that the pass order of all vehicles is consistent and non-conflicting, inter-group negotiation is conducted via roadside computing units, based on the intra-group pass order that has already been negotiated. The approach of first performing intra-group V2V cooperation to negotiate the intra-group pass order, followed by vehicle-to-infrastructure (V2I) cooperation for inter-group negotiation, helps avoid the issue of cyclic pass orders that may arise when directly solving for the global pass order with a large number of vehicles. It also reduces the computational complexity for roadside computing units.

To achieve this goal, the roadside computing system converts the key information of the scenario into a natural language prompt, which is then fed into the LLM. The prompt includes the following information: (1) the initial pass order of each vehicle group; (2) the conflict information for each pair of conflicting vehicles, including their speeds and distances to the conflict point; (3) the "leader-follower" car-following relationship for each pair of following vehicles.
The following requirements must also be met: (1) ensure that the leading vehicle in a car-following relationship passes first, while the following vehicle passes later; (2) no vehicles that appear in the intra-group pass order decision results should be omitted or newly added; (3) ensure that the priority relationships of vehicles within each group are maintained in the inter-group negotiation results, and no results that contradict the intra-group pass order should be generated.
The output format is a vehicle pass order list, in which the ID of each vehicle and its corresponding group are recorded sequentially.

\vspace{-0.3cm}
\subsection{Action Calculation}
To reduce the computational complexity of roadside computing devices, only the pass order is determined in the negotiation decision-making process. Therefore, based on the calculated pass order, an additional mechanism is designed to compute the control actions of individual autonomous vehicles.

After the pass order is determined, the acceleration calculation method for vehicles is centered on the core objectives of safety gap constraints and dynamic acceleration control. The goal is to ensure that each vehicle can safely pass through the conflict point while maximizing traffic efficiency. First, the system obtains the state information of each vehicle and performs initial processing to acquire key parameters such as the vehicle's current position, speed, and distance to the conflict point. For each vehicle, the system calculates its earliest possible passing time based on the passing time of its preceding vehicle and the preset safety time gap \( \Delta t_{\text{safe}} \). For each consecutive pair of vehicles \( (i, i+1) \), the system imposes the time constraint \( t_{i+1} \geq t_i + \Delta t_{\text{safe}} \). Based on this constraint, the system dynamically adjusts the acceleration of each vehicle in real-time to ensure safe passage.

\section{Numerical Experiment}
The feasibility and effectiveness of the proposed vehicle group division based on impact quantification and the LLM-based pass order negotiation framework was analyzed through numerical experiment using a designed case study. From the perspective of safety and efficiency, the simulation results of three methods, individual-vehicle decision-making (IVD), intra-group negotiation (IGN), and intra-group \& inter-group negotiation (I\&IGN) were analyzed under different vehicle number settings at the intersection.
\subsection{Case Design}
The experiment was configured using highway-env \cite{highway-env}, simulating a two-lane intersection scenario with 2, 4, and 8 vehicles. The entrance direction, initial position, speed and expected direction of vehicles were randomly generated. For each configuration, 10 random seeds were used to produce specific cases. Detailed experimental parameters were provided in Table ~\ref{Sim_params}.

\begin{table}[htbp]
    \vspace{-0.2cm}
    \caption{the model parameters of the designed case}
    \centering
    \begin{tabular}{c c c}
    \hline
     Symbol & Meaning & Value\\
     \hline

      ${d}_{0}$ & Initial Distance to Stop Line & 40.0-80.0 m \\
      ${v}_{0}$ & Initial Speed & 30.0-31.0 km/h \\
      ${seed}$ & Seed & 0-9 \\
      ${l}$ & Vehicle Length & 5.0m\\
      ${R_{IVD}}$ & Detect Range of IVD & 80.0m \\
     \hline
    \end{tabular}

    \label{Sim_params}
    \vspace{-0.2cm}
\end{table}

The Individual Vehicle Decision (IVD) involved each CAV in the scenario generating a reasonable inference of the intersection passage order for itself and surrounding vehicles, based on the state of vehicles within a certain distance from the ego vehicle which was inferred from the perspective of this CAV's safety and efficiency, and this vehicle planned its motion accordingly. Intra-Group Negotiation (IGN) referred to dividing all vehicles in the environment into several vehicle groups based on the impact quantification results, followed by negotiation within each group to determine the pass order. The motion of vehicles within each group was then planned based on the resulting intra-group order. In the case of Intra-Group and Inter-Group Negotiation (I\&IGN), after completing the intra-group negotiation, an additional inter-group negotiation was conducted to obtain a global pass order, as described in Section \uppercase\expandafter{\romannumeral3}.

\subsection{Safety and Efficiency Analysis}
\begin{figure}[htbp]
  \begin{center}
  \centerline{\includegraphics[width=3.5in]{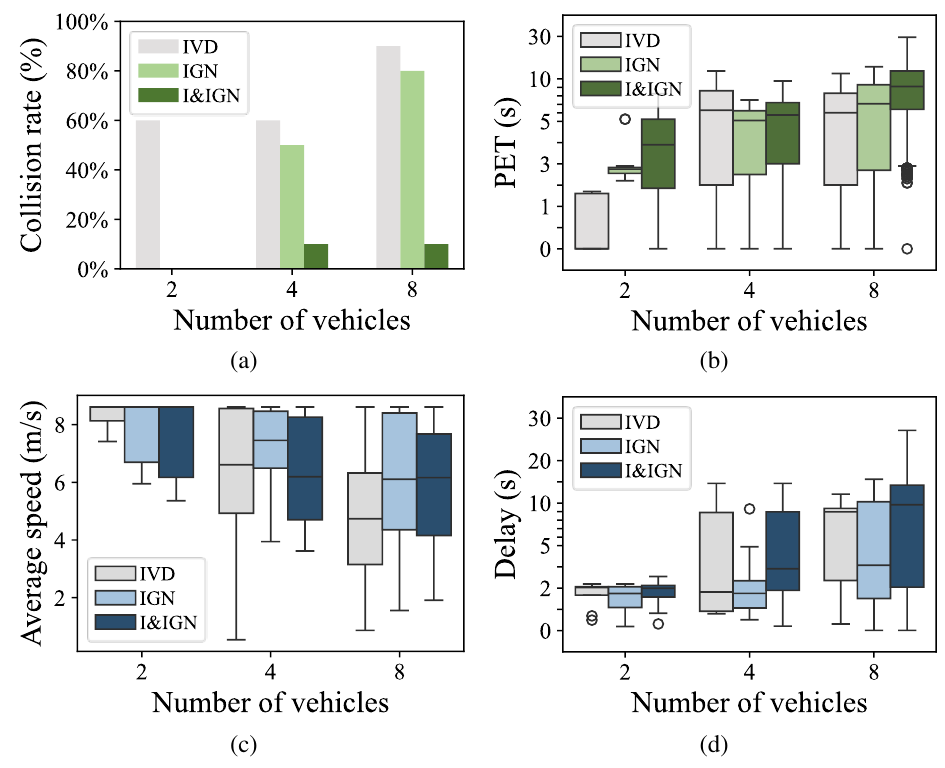}}
  \caption{Safety and efficiency results for different method.}\label{fig_sim_safeeff}
  \end{center}
  \vspace{-1.0cm}
\end{figure}

The simulation results were analyzed from safety and efficiency perspectives.
As the number of interacting vehicles increases, the complexity of the scenario rises. Severe conflicts occurred between vehicles with similar positions and speeds near the conflict zone. Consequently, the collision rate was highest when using the single-vehicle decision approach for all vehicle counts, as shown in Fig.~\ref{fig_sim_safeeff}(a). When introducing negotiation of pass order within vehicle groups (V2V-IGN), the collision rate decreased. However, in the more complex 8-vehicle scenario, negotiation only within the group was still insufficient to ensure safety. Our proposed V2I-I\&IGN system ensured safe passage in the majority of scenarios (2 vehicles-100\%, 4 vehicles-90\%, and 8 vehicles-90\%), particularly for more complex scenes with a higher number of vehicles, showing a significant improvement in safety.

The PET (Post-Encroachment Time) between each pair of vehicles in the scene was calculated and presented in Fig.~\ref{fig_sim_safeeff}(b). Compared to the IVD approach, the PET of the V2I-I\&IGN method was also improved, particularly in the 8-vehicle scenario.

The average speed and delay of all vehicles were selected as efficiency metrics for analysis shown in Fig.~\ref{fig_sim_safeeff}(c)(d). For all vehicle numbers, the speed and delay of different methods did not show significant differences or trends when observed as a whole, indicating that while the negotiation methods significantly improved safety, there was no notable loss in efficiency, thus effectively accomplishing intersection vehicle passage tasks.
\vspace{-0.3cm}

\subsection{LLM Negotiation-related Metrics}
\begin{figure}[htbp]  
\centering
\begin{minipage}[b]{0.24\textwidth}  
    \centering
    \includegraphics[width=\linewidth]{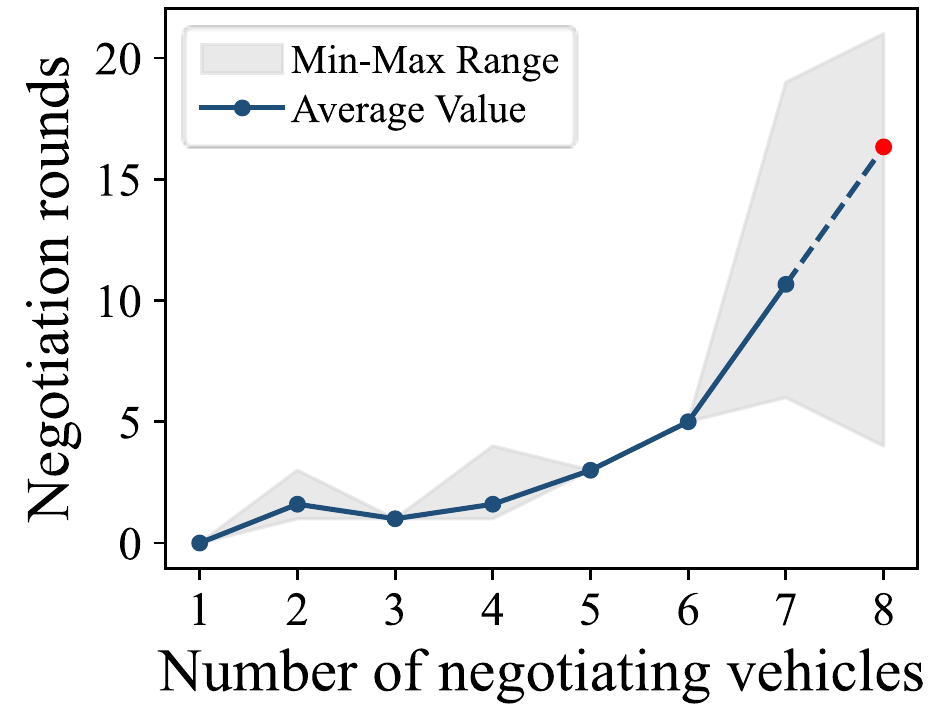}  
    \vspace{-0.5cm}
    \caption{Negotiation counts under different numbers of vehicles.}
    \label{fig_sim_negocounts}
\end{minipage}
\hfill  
\begin{minipage}[b]{0.24\textwidth}  
    \vspace{-0.8cm}
    \centering

    \begin{table}[H]  
        \caption{Negotiation rounds statistics}
        \vspace{-0.2cm}
        \centering
        \begin{tabular}{c| c c c}
        \hline
         \multirow{2}{*}{Veh num} & \multicolumn{3}{c}{Nego rounds}\\
         \cline{2-4}
           & Aver & Min & Max\\
         \hline
         1 & 0 & 0 & 0 \\
         2 & 1.6 & 1 & 3 \\
         3 & 1.0 & 1 & 1 \\
         4 & 1.6 & 1 & 4 \\
         5 & 3.0 & 3 & 3 \\
         6 & 5.0 & 5 & 5 \\
         7 & 10.7 & 6 & 19 \\
         8 & 16.3 & 4 & 21 \\
         \hline
        \end{tabular}
        \label{Sim_rounds_sta}
    \end{table}
\end{minipage}
\vspace{-0.4cm}
\end{figure}

From the perspective of the number of negotiation rounds required to reach a pass order, we analyzed the negotiation rounds under different vehicle counts and highlighted the necessity of vehicle group division. The number of negotiation rounds was counted for the 8-vehicle setting under different vehicle groupings, as well as for the case where no group division was performed and all vehicles negotiated collectively. The results, shown in Fig.~\ref{fig_sim_negocounts}, indicated that as the number of negotiating vehicles increased, the required negotiation rounds also increased, with a more significant rise occurring when the number of vehicles exceeded 6. Due to the imposed limit of a maximum of 20 renegotiation rounds, the maximum number of negotiation rounds in the figure was capped at 21. Without this limit, the negotiation rounds for the 8-vehicle case without group division would be higher, while other settings with fewer vehicles did not reach the maximum limit.

The specific number of negotiation rounds for each vehicle count was shown in Table.~\ref{Sim_rounds_sta}.

\vspace{-0.0cm}
\subsection{Detailed Explanation of Typical Case}
\begin{figure}[htbp]
  \begin{center}
  \centerline{\includegraphics[width=3.5in]{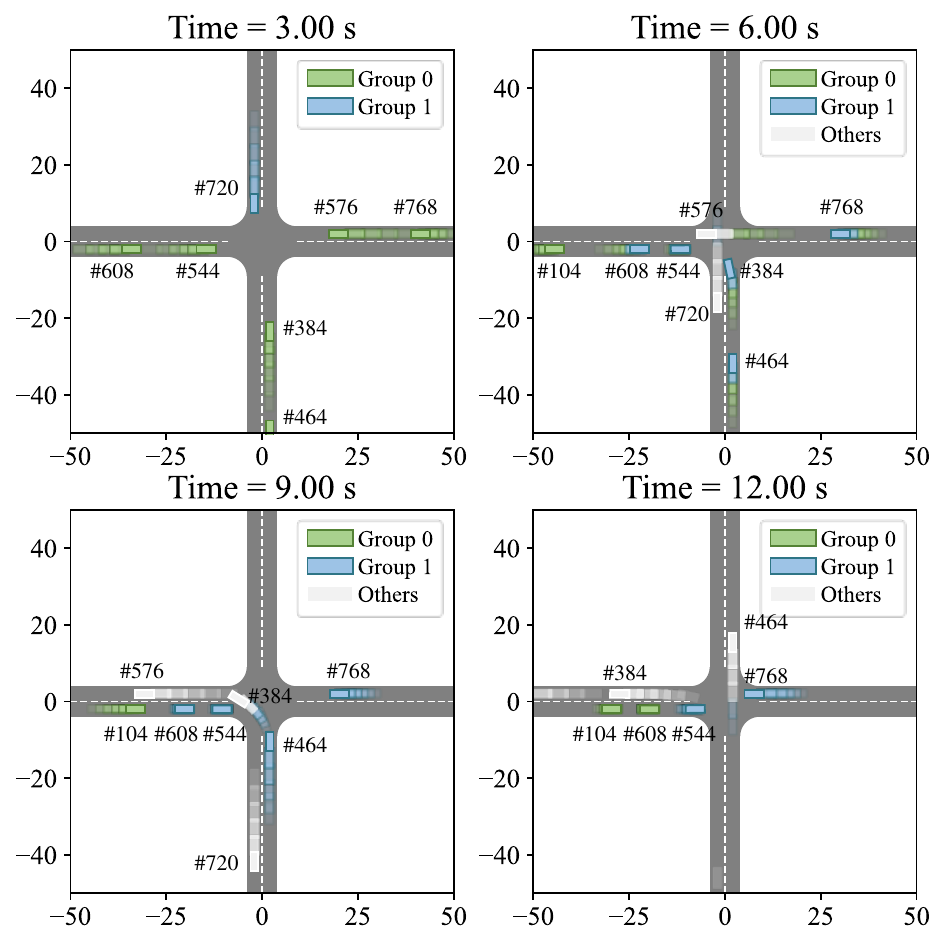}}
  \caption{Vehicle states of typical case during pass process.}\label{fig_sim_status}
  \end{center}
  \vspace{-1.0cm}
\end{figure}

A specific case with 8 vehicles was analyzed in detail, as shown in Fig.~\ref{fig_sim_status}. At \( t = 3 \, \text{s} \), vehicle \#720, being fast and closer to the conflict zone, should pass through the intersection first. It had minimal influence on other vehicles, so it was assigned to a separate group and proceeded at maximum speed. At \( t = 6 \, \text{s} \), vehicles \#720 and \#576 had passed through the intersection, and the vehicle with a higher interaction intensity (the blue vehicle) was assigned to a group. At \( t = 12 \, \text{s} \), as the vehicles passed through, the strong car-following influence between vehicles \#104 and \#608 led to their grouping as a single entity. Finally, vehicles \#768, \#544, \#608, and \#104 passed in sequence.
\section{Conclusion}
In this study, we propose a cooperative decision-making framework for CAVs at unsignalized intersections based on impact quantification and LLM-based communication sequence negotiation. This framework achieves an organic integration of V2V and V2I cooperation. First, the direct impact calculation method and impact propagation relationships between vehicles are defined, leading to the accumulation of vehicle impacts, which is then used for the division of vehicle groups with strong influences. Next, negotiation of the passing sequence is conducted both within and between vehicle groups based on LLM. Through numerical simulation analysis, the effectiveness of the vehicle grouping and negotiation method is validated, demonstrating its capability to safely and efficiently achieve intersection passage tasks.

This framework also provides avenues for future research. On one hand, unlike the current approach which only considers the impact relationships at the present moment, future research could explore the group division by incorporating temporal influence. On the other hand, considering the long-term coexistence of human-driven vehicles (HDVs) and AVs, future work could include integrating HDVs into the negotiation framework.
\ifCLASSOPTIONcaptionsoff
  \newpage
\fi

\footnotesize
\bibliographystyle{IEEEtranN}
\bibliography{IEEEabrv,Bibliography}

\vfill

\end{document}